\documentclass{article}
\usepackage{spconf,amsmath,graphicx}
\usepackage{float}

\usepackage{caption}
\usepackage{subfigure}
\usepackage{physics}
\usepackage{amsmath}
\usepackage{tikz}
\usepackage{yhmath}
\usepackage{cancel}
\usepackage{color}
\usepackage{siunitx}
\usepackage{array}
\usepackage{multirow}
\usepackage{amssymb}
\usepackage{gensymb}
\usepackage{tabularx}
\usepackage{extarrows}
\usepackage{booktabs}
\usepackage{pgfplots}
\usetikzlibrary{fadings}
\usetikzlibrary{patterns}
\usetikzlibrary{shadows.blur}
\usetikzlibrary{shapes}
\usepackage{colortbl}
\usepackage{xcolor}
\definecolor{lgreen}{rgb}{0.937,0.992,0.929}
\definecolor{dgreen}{rgb}{0.470,0.650,0.369}
\definecolor{lblue}{rgb}{0.902,0.933,1.000}
\definecolor{dblue}{rgb}{0.090,0.478,0.820}
\definecolor{dodgerblue}{rgb}{0.117,0.564,1.000}
\definecolor{orangered}{rgb}{1.000,0.270,0.000}
\definecolor{nasdaqup}{rgb}{0.000,0.654,0.356}
\usepgfplotslibrary{groupplots} 
\usepgfplotslibrary[groupplots]
\usetikzlibrary{pgfplots.groupplots}
\usetikzlibrary[pgfplots.groupplots]
\usetikzlibrary{fit}
\usetikzlibrary{backgrounds}

\ninept

\newcommand{\etal}{\emph{et al.}}

\title{Soft Alignment of Modality Space for End-to-end \\Speech Translation}

\name{Yuhao Zhang$^{1*}$ Kaiqi Kou$^{1*}$ Bei Li$^{1}$ Chen Xu$^{2}$ 
Chunliang Zhang$^{1,3}$ Tong Xiao$^{1,3\dagger}$ Jingbo Zhu$^{1,3}$
\thanks{
* Equal contribution.
}
\thanks{$\dagger$ Corresponding author. }
}

\address{$^{1}$ School of Computer Science and Engineering, Northeastern University, Shenyang, China \\
$^{2}$ Harbin Engineering University \\
$^{3}$ NiuTrans Research, Shenyang, China \\
}

\begin{document}

\maketitle

\begin{abstract}

End-to-end Speech Translation (ST) aims to convert speech into target text within a unified model. The inherent differences between speech and text modalities often impede effective cross-modal and cross-lingual transfer. Existing methods typically employ hard alignment (H-Align) of individual speech and text segments, which can degrade textual representations. To address this, we introduce Soft Alignment (S-Align), using adversarial training to align the representation spaces of both modalities. S-Align creates a modality-invariant space while preserving individual modality quality. Experiments on three languages from the MuST-C dataset show S-Align outperforms H-Align across multiple tasks and offers translation capabilities on par with specialized translation models.
\end{abstract}


\section{Introduction}
\label{sec:intro}

The End-to-end speech translation (ST) model has attracted many researchers attentions due to its low latency and prevention of error propagation compared to the cascade model \cite{berard2016listen, duong2016attentional}. It directly translates speech input to target language text, necessitating simultaneous cross-modal and cross-lingual capabilities. However, the scarcity of ST training data makes building the model challenging without additional training data and strategies \cite{dong2021listen}. Consequently, researchers are increasingly adopting the pre-training approach \cite{li-etal-2021-multilingual, zhang-etal-2022-speechut, zhang-etal-2023-rethinking}. This involves initially pre-training ST model modules with auxiliary tasks, equipping them with foundational cross-modal and cross-lingual abilities, followed by overall fine-tuning for the ST task \cite{ye21_interspeech, zheng2021fused}. This strategy effectively enhances the performance of end-to-end model, and even outperforms the cascade model when dealing with limited ST training data \cite{xu-etal-2021-stacked}.

\begin{figure}[t]
  \centering
    \begin{tikzpicture}
    \node(fbanks) at(0,0) {\includegraphics[width=2.5cm,height=0.8cm] {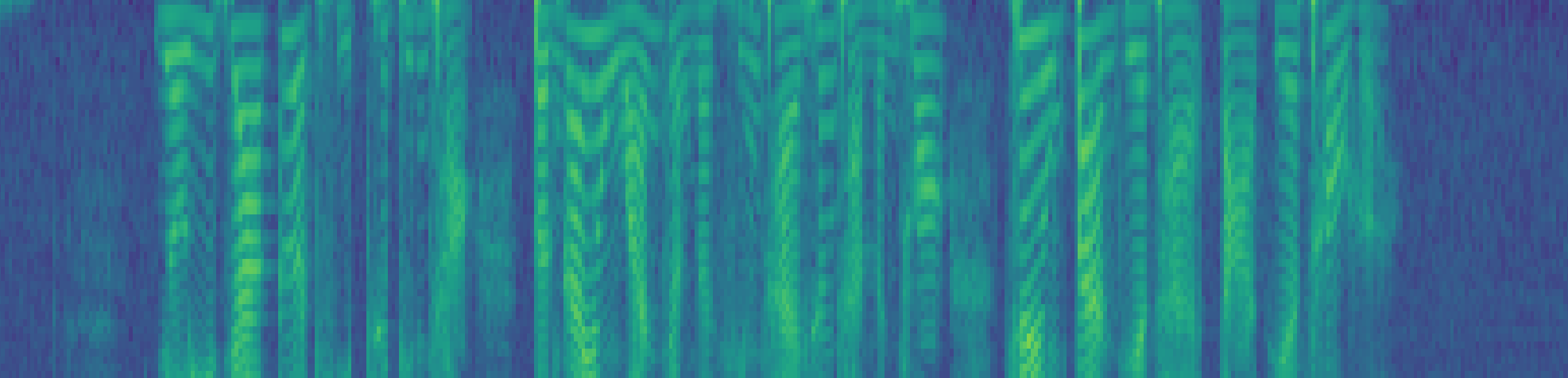}};
    
    \node(text) at ([xshift=2.2cm]fbanks.east) {How are you};
    \scalebox{1.1}{
    \filldraw[shift={(-9pt,21pt)}, fill=lgreen, smooth cycle, draw=none]
    plot coordinates {
      (-3pt,-3pt) (-8pt,15pt) (-8pt,30pt) (-3pt,38pt) (3pt,44pt) (12pt,48pt) (28pt,47pt) (32pt,38pt) (31pt,28pt) (22pt,6pt) (12pt,-2pt)
    } ;}
    \scalebox{1.2}{
    \filldraw[shift={(81pt,15pt)}, fill=lblue, smooth cycle, draw=none]
    plot coordinates {
      (0pt,0pt) (-20pt,25pt)  (-17pt,40pt) (-7pt,47pt) (15pt,40pt) (25pt,20pt) (15pt,3pt)
      } ;}
      
    \node[align=center, fill=dgreen, inner sep=0pt, minimum width=5pt, minimum height=5pt](s1) at ([xshift=-3pt, yshift=48pt]fbanks.north) {};
    \node[ isosceles triangle, align=center, fill=dgreen, inner sep=0pt, rotate=-30, isosceles triangle apex angle=60, minimum width=5pt, minimum height=5pt](s2) at ([xshift=-6pt, yshift=27pt]fbanks.north) {};
    \node[ circle, align=center, fill=dgreen, inner sep=0pt, minimum width=5pt, minimum height=5pt](s3) at ([xshift=5pt, yshift=38pt]fbanks.north) {};
    
    \node[align=center, fill=dblue, inner sep=0pt, minimum width=5pt, minimum height=5pt](t1) at ([xshift=-14pt, yshift=40pt]text.north) {};
    \node[isosceles triangle, align=center, fill=dblue, rotate=-30, isosceles triangle apex angle=60, inner sep=0pt, minimum width=5pt, minimum height=5pt](t2) at ([xshift=5pt, yshift=28pt]text.north) {};
    \node[circle, align=center, fill=dblue, inner sep=0pt, minimum width=5pt, minimum height=5pt](t3) at ([xshift=0pt, yshift=48pt]text.north) {};

    \draw[->,thick,dgreen] ([xshift=-0.7cm,yshift=-0.1cm]fbanks.north)..controls([xshift=-0.8cm,yshift=-1.2cm]s1.south) and ([xshift=-0.8cm,yshift=-0.4cm]s1.south)..(s1.west);
    \draw[->,thick,dgreen] ([xshift=-0.1cm,yshift=-0.1cm]fbanks.north)..controls([xshift=-7pt,yshift=8pt]fbanks.north) and ([xshift=-7pt,yshift=16pt]fbanks.north) ..([xshift=1pt]s2.south); 
    \draw[->,thick,dgreen] ([xshift=0.6cm,yshift=-0.1cm]fbanks.north)..controls([xshift=24pt,yshift=14pt]fbanks.north) and ([xshift=18pt,yshift=30pt]fbanks.north) ..([xshift=-1pt,yshift=-1pt]s3.east); 

    \draw[->,thick,dblue] ([xshift=-0.7cm,yshift=-0.1cm]text.north)..controls([xshift=-28pt,yshift=10pt]text.north) and ([xshift=-28pt,yshift=30pt]text.north)..(t1.west);
    \draw[->,thick,dblue] ([xshift=0cm,yshift=-0.1cm]text.north)..controls([xshift=-2pt,yshift=6pt]text.north) and ([xshift=-2pt,yshift=16pt]text.north) ..([xshift=1pt]t2.south); 
    \draw[->,thick,dblue] ([xshift=0.7cm,yshift=-0.1cm]text.north)..controls([xshift=24pt,yshift=14pt]text.north) and ([xshift=24pt,yshift=34pt]text.north) ..([xshift=-1pt,yshift=-1pt]t3.east);
     
    \draw[<->,dashed, very thick,red!50] ([xshift=0pt,yshift=0pt]s1.north) ..controls([xshift=25pt,yshift=15pt]s1.north) and ([xshift=80pt,yshift=13pt]s1.north) ..(t1.north);
    \node[align=center](hard) at ([xshift=50pt,yshift=-3pt]s1.east) {\scriptsize{Hard alignment}};
     
    \draw[<->,dashed, very thick,red!50] (3pt,75pt) ..controls(32pt,92pt) and (82pt,82pt) .. (98pt,73pt);
    \node[align=center](hard) at (50pt,90pt) {\scriptsize{Soft alignment}};
    
    \node[align=left](speech_space) at ([xshift=-10pt,yshift=60pt]fbanks.west) {\scriptsize{Speech}\\ \scriptsize{modality}\\ \scriptsize{space}};
    \node[align=right](text_space) at ([xshift=10pt,yshift=60pt]text.east) {\scriptsize{Text}\\ \scriptsize{modality}\\ \scriptsize{space}};
    \end{tikzpicture}
  
    \caption{The comparison of two alignment methods. The hard strategy aims to align each speech and text while soft method aligns the representation space of two modalities.}
    \label{soft_alignment}
    
\end{figure}
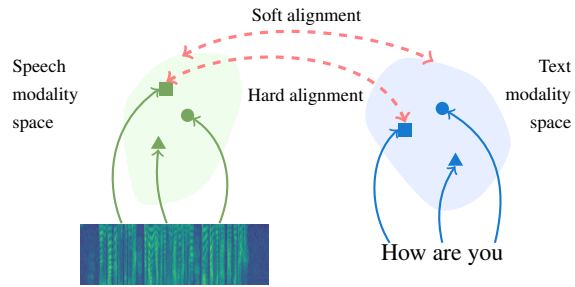

However, pre-training methods struggle with the modality gap between speech and text representation \cite{zhang2022improving}. Speech features, derived from signal time variation, often include noise and blanks, contrasting the cleaner single-word information in text features. Additionally, speech sequences are significantly longer than text \cite{zhang23u_interspeech, xu-etal-2023-bridging}, hindering effective pre-training and unsuitable for pre-trained textual tasks. Some researchers use alignment loss like contrastive learning \cite{ye-etal-2022-cross, ouyang-etal-2023-waco} to address this issue, aiming to align speech and text as shown in Figure \ref{soft_alignment} (referred to as hard alignment). While effective in bridging the gap for speech translation (ST) tasks, Table \ref{hard_alignment} demonstrates that this approach harms machine translation (MT) performance, with heightened contrastive learning exacerbating the issue. This shows the intrinsic modal differences preventing further mitigation of the modality gap through hard alignment (H-Align).

\begin{table}[t] 
\centering
\small
\setlength{\tabcolsep}{3mm}{
\begin{tabular}{lllcccc}
\toprule
&\ \ ST&\ MT \\
\hline

Single model&27.9  & 33.7   \\
Hard alignment & 28.3(\textcolor{nasdaqup}{$+$0.4}) & 30.5 (\textcolor{red}{$-$3.2})  \\
\bottomrule
\end{tabular}
}
\caption{Results on MuST-C En-De with external MT data. \textit{Single model} denotes the model only achieves one task. We implement the contrastive learning as hard alignment.} 
\label{hard_alignment}
\end{table}

To address this drawback, we introduce the soft alignment (S-Align) approach, which aims to align the representation space rather than individual sample pairs, as depicted in Figure \ref{soft_alignment}. Our method involves a modal network that classifies representations by modality, employing adversarial training (AT) \cite{Goodfellow_gan, ganin2016domain} to optimize the model. When the modal network struggles to identify a representation's modality, it fosters representation space alignment. Additionally, we propose the continuity method to convert discrete predictions into a continuous space, further enhancing the impact of soft alignment. This technique effectively bridges the modality gap, preserving task representation space and mitigating losses across tasks.

Experimental results on the MuST-C dataset, demonstrating the effectiveness of our approach with significant improvements over a strong baseline. Moreover, our method accomplishes MT, ASR, and ST decoding tasks within a unified model, preventing the waste of pre-training resources. Our results consistently surpass those of individual models, underscoring the successful alignment of representations achieved by our approach\footnote{Our code is available at https://github.com/MuKai2000/S-Align.}.

\section{Method}
\label{sec:method}

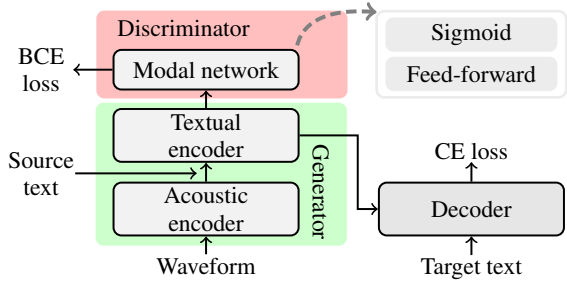
\begin{figure}[t]
\centering
\tikzset{every picture/.style={line width=0.75pt}}   

\begin{tikzpicture}
    \node[draw, rounded corners=3pt, align=center, fill=gray!10, inner sep=0pt, minimum width=70pt, minimum height=20pt](aencoder) at (0pt,0pt) {Acoustic \\encoder};
    \node[draw, rounded corners=3pt, align=center, fill=gray!10, inner sep=0pt, minimum width=70pt, minimum height=20pt](tencoder) at ([xshift=0pt, yshift=17.5pt]aencoder.north) {Textual \\encoder};
    \node[draw, rounded corners=3pt, align=center, fill=gray!20, inner sep=0pt, minimum width=70pt, minimum height=20pt](decoder) at ([xshift=65pt, yshift=0pt]aencoder.east) {Decoder};
    \node[draw, rounded corners=3pt, align=center, fill=gray!10, inner sep=0pt, minimum width=70pt, minimum height=15pt](tasknet) at ([xshift=0pt, yshift=14.5pt]tencoder.north) {Modal network};
    \node[rounded corners=3pt, align=center, fill=gray!15, inner sep=0pt, minimum width=65pt, minimum height=13pt](FFN) at ([xshift=65pt, yshift=-1.6pt]tasknet.east) {Feed-forward};
    \node[rounded corners=3pt, align=center, fill=gray!15, inner sep=0pt, minimum width=65pt, minimum height=13pt](sigmoid) at ([xshift=0pt, yshift=8pt]FFN.north) {Sigmoid};

    \node[align=center, inner sep=0pt, rotate=-90](generator) at ([xshift=8pt, yshift=6.5pt]aencoder.east) {Generator};
    \node[align=center, inner sep=0pt](discriminator) at ([xshift=-8.5pt, yshift=8pt]tasknet.north) {Discriminator};

    \node[align=center](wave) at ([xshift=0pt, yshift=-11.25pt]aencoder.south){Waveform};
    \node[align=center](src) at ([xshift=-27pt, yshift=13.5pt]aencoder.west){Source\\text};
    \node[align=center](tgt) at ([xshift=0pt, yshift=-12pt]decoder.south){Target text};
    \node[align=center](ce) at ([xshift=0pt, yshift=11.5pt]decoder.north){CE loss};
    \node[align=center](bce) at ([xshift=-27pt, yshift=0pt]tasknet.west){BCE\\loss};

    \draw[->] ([xshift=0pt, yshift=-2pt]wave.north) -- (aencoder.south);
    \draw[->] ([xshift=-3pt, yshift=0pt]src.east) -- ([xshift=43pt, yshift=0pt]src.east);
    \draw[->] (aencoder.north) -- (tencoder.south);
    \draw[->] (tencoder.north) -- (tasknet.south);
    \draw[->] (tencoder.east) -- ([xshift=22pt, yshift=0pt]tencoder.east) -- ([xshift=22pt, yshift=0pt]aencoder.east) -- (decoder.west);

    \draw[->] ([xshift=0pt, yshift=-2.4pt]tgt.north) -- (decoder.south);
    \draw[->] (decoder.north) -- ([xshift=0pt, yshift=2.4pt]ce.south);
    \draw[->] (tasknet.west) -- (bce.east);

    \draw[->, dashed, dash pattern=on 4pt off 2pt, color=gray, line width=1.5pt] ([xshift=-0.9pt, yshift=9.2pt]tasknet.east) .. controls ([xshift=0pt, yshift=13.2pt]tasknet.east) and ([xshift=-28pt, yshift=10.5pt]sigmoid.west) .. ([xshift=-3pt, yshift=8.5pt]sigmoid.west);

    \begin{pgfonlayer}{background}
        \node[rounded corners=3pt, fill=green!20, fit=(aencoder)(tencoder)(generator), inner xsep=6pt, inner ysep=2pt, minimum width=95pt] {};
        \node[rounded corners=3pt, fill=red!25, fit=(tasknet)(discriminator), inner xsep=6pt, inner ysep=3pt, minimum width=95pt, minimum height=25pt, shift={(6pt,0pt)}] {};
        \node[draw=gray!15, line width=1pt, rounded corners=3pt, fit=(FFN)(sigmoid), inner xsep=3pt, inner ysep=2pt, minimum width=70pt, minimum height=30pt] {};
    \end{pgfonlayer}
    
\end{tikzpicture}
\caption{The overall architecture of our model. It is a variant based on generative adversarial networks.}
\label{fig:at_architecture}
\end{figure}

\subsection{Overview of Architecture}
Our model shown in Figure \ref{fig:at_architecture} follows an encoder-decoder architecture with four key modules: acoustic encoder (A-enc), textual encoder (T-enc), decoder, and modal network. A-enc takes audio waveforms as input, converting them into token-level representations, usually pre-trained through self-supervised learning \cite{baevski2020wav2vec}. T-enc utilizes either source text or A-enc output, extracting semantic information for cross-lingual transfer. The decoder performs autoregressive translation based on the given source. The entire architecture employs the Transformer \cite{vaswani2017attention} as the backbone network.

We narrow the modality gap through adversarial training in the modality space. Our encoders, acting as generators in the adversarial setup, differ T-enc for text and all encoders for speech. A classifier serves as the discriminator, featuring a feed-forward network and an output layer for binary modality prediction. The sigmoid function handles classification due to the two modalities. We incorporate average pooling to adjust adversarial training to sentence level.

\begin{table*}[t] 
\centering
\small
\setlength{\tabcolsep}{4mm}{
\begin{tabular}{l|ccccccll}
\toprule
\multirow{2}{*}{Model}&\multicolumn{2}{c}{En-De}&\multicolumn{2}{c}{En-Fr}&\multicolumn{2}{c}{En-Es}&\multicolumn{2}{c}{Avg.}\\
&ST&MT&ST&MT&ST&MT&\multicolumn{1}{c}{ST}&\multicolumn{1}{c}{MT} \\ \hline
\multicolumn{9}{c}{Pre-train $w/o$ external MT data}\\ \hline
ConST \cite{ye-etal-2022-cross}&25.7 & -&36.8&-&30.4&-& 31.0&- \\
STEMM \cite{fang-etal-2022-stemm}&25.6 & -&36.1&-&30.3&-&30.7&- \\
\hline
Single model&24.3&30.5& 35.2&42.7&29.6&35.5&29.7&36.2\\

\rowcolor{black!10} H-Align&26.3&29.2&37.5&39.5&31.2&34.8&31.7 (\textcolor{nasdaqup}{$+$2.0})&34.5 (\textcolor{red}{$-$1.7})\\
\rowcolor{black!10} S-Align&\textbf{26.5}&\textbf{30.9}&\textbf{37.6}&\textbf{42.9}&\textbf{31.3}&\textbf{35.8}&\textbf{31.8} (\textcolor{nasdaqup}{$+$2.1})&\textbf{36.5} (\textcolor{nasdaqup}{$+$0.3})\\

\hline
\multicolumn{9}{c}{Pre-train $w/$ external MT data}\\ \hline
XSTNet \cite{ye21_interspeech}&27.8&33.2 &38.0& 45.3&30.8&37.2$^{*}$&32.2&38.6 \\ 
ConST \cite{ye-etal-2022-cross}& 28.3&-&38.3&-& 32.0&-&32.9&-\\
\rowcolor{black!10} STEMM \cite{fang-etal-2022-stemm} &\textbf{28.7}&31.5&37.4&-&31.0&-&32.4&-  \\
\rowcolor{black!10} JT$^{\dag}$ \cite{tang-etal-2021-improving} & 26.8&30.5&37.4&42.3&31.0&34.7&31.7&35.8 \\
\hline
Single model&27.9&33.7& 36.8&\textbf{45.7}&30.2 &38.0&31.6&39.1 \\
\rowcolor{black!10} H-Align&28.3&30.5&39.8&36.1&32.0&37.9&33.4 (\textcolor{nasdaqup}{$+$1.8})&34.8 (\textcolor{red}{$-$4.3}) \\
\rowcolor{black!10} S-Align&28.6&\textbf{33.8}&\textbf{39.9}&\textbf{45.7}&\textbf{32.6}&\textbf{38.2}&\textbf{33.7} (\textcolor{nasdaqup}{$+$2.1})&\textbf{39.2} (\textcolor{nasdaqup}{$+$0.1}) \\

\bottomrule
\end{tabular}
}
\caption{Results on MuST-C. Grey background denotes the model can address two tasks while the others need fine-tune on the specific task. $\dag$ indicates that the model uses the Mel fbanks as the input and without the pre-trained acoustic encoder.  * means the reproduction result. }
\label{results}
\end{table*}

\subsection{Training Objective}

We enhance our baseline using the progressive training strategy \cite{ye21_interspeech}. Initially, we pre-train T-enc and decoder with the MT task, then employ multi-task learning to fine-tune the ST task. Specifically, we focus on improving the fine-tuning stage through multi-task learning in this study. The ST training set is denoted as ${(\mathbf{s}, \mathbf{x}, \mathbf{y})}$, where $\mathbf{s}$, $\mathbf{x}$, and $\mathbf{y}$ represent speech, transcription, and translation sequences, respectively. Furthermore, ${(\mathbf{s}, \mathbf{x})}$ and ${(\mathbf{x}, \mathbf{y})}$ refer to the ASR and MT training sets, respectively.

We introduce the ASR task to improve the ability of modeling speech and alleviate the pressure of cross-modal for T-enc \cite{le2023pre}. We use the CTC \cite{graves2006connectionist} loss to training ASR task:
\begin{equation}
    \mathcal{L}_{\mathrm{ASR}} = - \mathrm{log} \mathrm{P}_{\mathrm{CTC}}(\mathbf{x}|\mathbf{s})
\end{equation}
the CTC probability can be calculated as:
\begin{equation}
    \mathrm{P}_{\mathrm{CTC}}(\mathbf{x}|\mathbf{s}) = \sum_{\mathbf{\pi} \in \mathcal{B}^{-1}(\mathbf{x})}\mathrm{P}(\mathbf{\pi}|\mathbf{s})
\end{equation}
where $\mathcal{B}$ maps an alignment sequence $\mathbf{\pi}$ to $\mathbf{x}$.
We keep the MT task to avoid the catastrophic forgetting problem. The training objectives of MT and ST tasks are auto-regressive training as following:
\begin{equation}
    \mathcal{L}_{\mathrm{MT}} = -\sum_{i}^{|\mathbf{y}|} \mathrm{log} \mathrm{P}{(y_{i}|\mathbf{x}, y_{1:i-1})}
\end{equation}
\begin{equation}
    \mathcal{L}_{\mathrm{ST}} = -\sum_{i}^{|\mathbf{y}|} \mathrm{log} \mathrm{P}{(y_{i}|\mathbf{s}, y_{1:i-1})}
\end{equation}
where the $|\cdot|$ denotes the length of a sequence. We propose the adversarial training (AT) toward the modality representation. For the speech input $\mathbf{s}$, the output of the Generator $\mathcal{G}_{\mathrm{st}}$ can be the result of A-enc and T-enc, called $h_{\mathrm{st}} \in \mathbb{R}^{T \times D}$. And the $h_{\mathrm{mt}}$ denotes the output of $\mathcal{G}_{\mathrm{mt}}$, which has been processed by T-enc for each source text input. The goal of modal classifier, namely the Discriminator $\mathcal{D}$, is to identify the modality of input representation accurately. Thus the training objective can be described as follows: 
\begin{equation}
    \mathcal{L}_{\mathcal{D}} = - \mathrm{log} (\mathrm{P}{(c_\mathrm{st}|h_{\mathrm{st}}, \theta_{\mathcal{D}})}) - \mathrm{log} (\mathrm{P}{(c_\mathrm{mt}|h_{\mathrm{mt}}, \theta_{\mathcal{D}})})
\end{equation}
where the $c_{\mathrm{st}}$ and $c_{\mathrm{mt}}$ indicate the MT and ST category id (e.g. 0 and 1) separately. Then adversarial training requires the $\mathcal{G}_{\mathrm{st}}$ and $\mathcal{G}_{\mathrm{mt}}$ to cheat the $\mathcal{D}$, and our purpose is to learn a unified representation space, so we design the $c_{\mathrm{u}}$ as its label, thus the objectives are:
\begin{equation}
    \mathcal{L}_{\mathcal{G}_{\mathrm{st}}} = - \mathrm{log} (\mathrm{P}{(c_\mathrm{u}|h_{\mathrm{st}}, \theta_{\mathcal{D}})}) 
\end{equation}
\begin{equation}
    \mathcal{L}_{\mathcal{G}_{\mathrm{mt}}} = - \mathrm{log} (\mathrm{P}{(c_\mathrm{u}|h_{\mathrm{mt}}, \theta_{\mathcal{D}})})
\end{equation}
and we set the $c_{\mathrm{u}}$ to the average of $c_{\mathrm{st}}$ and $c_{\mathrm{mt}}$. The total training objective can represent as following:
\begin{equation}
\mathcal{L} = \mathcal{L}_{\mathrm{ASR}} +  \mathcal{L}_{\mathrm{MT}} + \mathcal{L}_{\mathrm{ST}} + \lambda (\mathcal{L}_{\mathcal{D}} + \mathcal{L}_{\mathcal{G}_{\mathrm{st}}} + \mathcal{L}_{\mathcal{G}_{\mathrm{mt}}})
\end{equation}
We can use the cross entropy (CE) loss for the $\mathcal{L}_{\mathrm{ST}}$ and $\mathcal{L}_{\mathrm{MT}}$. For adversarial training, we utilize the binary cross entropy (BCE) loss for $\mathcal{L}_{\mathcal{D}}$,  $\mathcal{L}_{\mathcal{G}_{\mathrm{st}}}$ and $\mathcal{L}_{\mathcal{G}_{\mathrm{mt}}}$ to achieve modality 
classification.

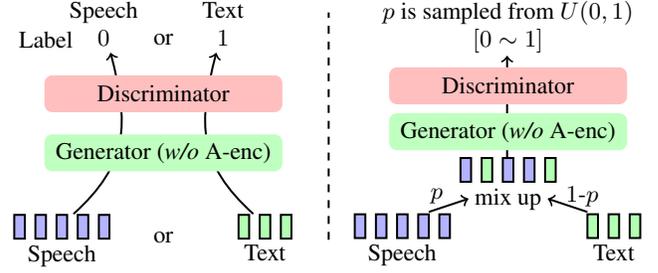
\begin{figure}[t]
\centering
\tikzset{every picture/.style={line width=0.75pt}} 

\begin{tikzpicture}

    \node[inner sep=0.3](or0) at (0pt ,0pt) {or};
    
    \node[align=center](text1) at ([xshift=34pt, yshift=-6pt]or0.east) {Text};
    \draw [draw=none] ([xshift=-18pt, yshift=8pt]text1.north) rectangle ([xshift=-14pt, yshift=-1pt]text1.north);
    \draw[fill=green!30] ([xshift=-10pt, yshift=8pt]text1.north) rectangle ([xshift=-6pt, yshift=-1pt]text1.north);
    \draw [fill=green!30] ([xshift=-2pt, yshift=8pt]text1.north) rectangle ([xshift=2pt, yshift=-1pt]text1.north);
    \draw [fill=green!30] ([xshift=6pt, yshift=8pt]text1.north) rectangle ([xshift=10pt, yshift=-1pt]text1.north);
    \draw [draw=none] ([xshift=14pt, yshift=8pt]text1.north) rectangle ([xshift=18pt, yshift=-1pt]text1.north);

    \node[align=center](speech1) at ([xshift=-34pt, yshift=-7pt]or0.west) {Speech};
    \draw [fill=blue!30] ([xshift=-18pt, yshift=8pt]speech1.north) rectangle ([xshift=-14pt, yshift=-1pt]speech1.north);
    \draw[fill=blue!30] ([xshift=-10pt, yshift=8pt]speech1.north) rectangle ([xshift=-6pt, yshift=-1pt]speech1.north);
    \draw [fill=blue!30] ([xshift=-2pt, yshift=8pt]speech1.north) rectangle ([xshift=2pt, yshift=-1pt]speech1.north);
    \draw [fill=blue!30] ([xshift=6pt, yshift=8pt]speech1.north) rectangle ([xshift=10pt, yshift=-1pt]speech1.north);
    \draw [fill=blue!30] ([xshift=14pt, yshift=8pt]speech1.north) rectangle ([xshift=18pt, yshift=-1pt]speech1.north);

    \draw[->, color=black] ([xshift=-34pt, yshift=6pt]or0.north) .. controls ([xshift=-13pt, yshift=25pt]or0.north) and ([xshift=-13pt, yshift=40pt]or0.north) .. ([xshift=-20pt, yshift=67pt]or0.north);
    \draw[->, color=black] ([xshift=34pt, yshift=6pt]or0.north) .. controls ([xshift=13pt, yshift=25pt]or0.north) and ([xshift=13pt, yshift=40pt]or0.north) .. ([xshift=20pt, yshift=67pt]or0.north);

    \node[rounded corners=3pt, align=center, fill=red!25, inner sep=0pt, minimum width=90pt, minimum height=14pt](dis1) at ([xshift=0pt, yshift=51.5pt]or0.north) {Discriminator};

    \node[rounded corners=3pt, align=center, fill=green!25, inner sep=0pt, minimum width=90pt, minimum height=14pt](gen1) at ([xshift=0pt, yshift=-15pt]dis1.south) {Generator (\textit{w/o} A-enc)};

    \node[inner sep=0.3](or1) at ([xshift=0pt, yshift=72pt]or0.north) {or};
    \node[inner sep=0.3, align=center] at ([xshift=-40pt]or1.west) {Label};
    
    \node[align=center] at ([xshift=27pt, yshift=6pt]or1.west) {Text\\$1$};
    \node[align=center] at ([xshift=-27pt, yshift=6pt]or1.east) {Speech\\$0$};

    \draw[->, color=black] ([xshift=130pt, yshift=-18pt]dis1.south) -- ([xshift=130pt, yshift=7pt]dis1.north);
    
    \draw[dashed, color=black] ([xshift=62.5pt, yshift=-73pt]or1.south) -- ([xshift=62.5pt, yshift=7.5pt]or1.north);

    \node[rounded corners=3pt, align=center, fill=red!25, inner sep=0pt, minimum width=90pt, minimum height=14pt](dis2) at ([xshift=85pt, yshift=3pt]dis1.east) {Discriminator};
    \node[rounded corners=3pt, align=center, fill=green!25, inner sep=0pt, minimum width=90pt, minimum height=14pt](gen2) at ([xshift=0pt, yshift=-10pt]dis2.south) {Generator (\textit{w/o} A-enc)};

    \node[align=center](text2) at ([xshift=40pt, yshift=-38.3pt]gen2.south) {Text};
    \draw [draw=none] ([xshift=-18pt, yshift=8pt]text2.north) rectangle ([xshift=-14pt, yshift=-1pt]text2.north);
    \draw[fill=green!30] ([xshift=-10pt, yshift=8pt]text2.north) rectangle ([xshift=-6pt, yshift=-1pt]text2.north);
    \draw [fill=green!30] ([xshift=-2pt, yshift=8pt]text2.north) rectangle ([xshift=2pt, yshift=-1pt]text2.north);
    \draw [fill=green!30] ([xshift=6pt, yshift=8pt]text2.north) rectangle ([xshift=10pt, yshift=-1pt]text2.north);

    \node[align=center](speech2) at ([xshift=-40pt, yshift=-39.3pt]gen2.south) {Speech};
    \draw [fill=blue!30] ([xshift=-18pt, yshift=8pt]speech2.north) rectangle ([xshift=-14pt, yshift=-1pt]speech2.north);
    \draw[fill=blue!30] ([xshift=-10pt, yshift=8pt]speech2.north) rectangle ([xshift=-6pt, yshift=-1pt]speech2.north);
    \draw [fill=blue!30] ([xshift=-2pt, yshift=8pt]speech2.north) rectangle ([xshift=2pt, yshift=-1pt]speech2.north);
    \draw [fill=blue!30] ([xshift=6pt, yshift=8pt]speech2.north) rectangle ([xshift=10pt, yshift=-1pt]speech2.north);
    \draw [fill=blue!30] ([xshift=14pt, yshift=8pt]speech2.north) rectangle ([xshift=18pt, yshift=-1pt]speech2.north);

    \node[ align=center](mix) at ([xshift=0pt, yshift=-18pt]gen2.south) {mix up};
    \draw [fill=blue!30] ([xshift=-18pt, yshift=8pt]mix.north) rectangle ([xshift=-14pt, yshift=-1pt]mix.north);
    \draw[fill=green!30] ([xshift=-10pt, yshift=8pt]mix.north) rectangle ([xshift=-6pt, yshift=-1pt]mix.north);
    \draw [fill=blue!30] ([xshift=-2pt, yshift=8pt]mix.north) rectangle ([xshift=2pt, yshift=-1pt]mix.north);
    \draw [fill=blue!30] ([xshift=6pt, yshift=8pt]mix.north) rectangle ([xshift=10pt, yshift=-1pt]mix.north);
    \draw [fill=green!30] ([xshift=14pt, yshift=8pt]mix.north) rectangle ([xshift=18pt, yshift=-1pt]mix.north);

    \draw[->, color=black] ([xshift=-30pt, yshift=-23pt]gen2.south) -- ([xshift=-15pt, yshift=-18pt]gen2.south);
    \draw[->, color=black] ([xshift=30pt, yshift=-23pt]gen2.south) -- ([xshift=15pt, yshift=-18]gen2.south);

    \node[inner sep=0.3, align=center](P) at ([xshift=0pt, yshift=15pt]dis2.north) {$p$ is sampled from $U(0,1)$\\ $[0\sim1]$};

    \node[inner sep=0.3, align=center] at ([xshift=-27pt, yshift=-17pt]gen2.south) {$p$};
    \node[inner sep=0.3, align=center] at ([xshift=28pt, yshift=-17pt]gen2.south) {$1$-$p$};
    
\end{tikzpicture}
\caption{The architecture of adversarial training. The left is
discrete prediction space and the right is continuous prediction space. Note that the speech tokens are outputs of the acoustic encoder.}
\label{fig:mixup}
\end{figure}

\subsection{Enhanced Adversarial Training}

In standard adversarial training, classifier $\mathcal{D}$ predicts within a discrete space, $\{c_\mathrm{st}, c_\mathrm{mt}\}$. Due to significant ST and MT sequence distinctions, $\mathcal{D}$ can classify input representation precisely, making it difficult for generators to achieve modality agnosticism. Model updating encounters the vanishing gradient issue with overly simple prediction space, hindering sufficient information for updates and preventing modality alignment. To address this, we introduce enhanced adversarial training as Figure \ref{fig:mixup} shown.

Our primary approach for achieving a continuous prediction space is mix-up method \cite{verma2019manifold}. Initially, we establish a uniform distribution $U(c_\mathrm{st}, c_\mathrm{mt})$, represented as $U(0,1)$ here, and then sample $p$ as the mix-up rate. The parameter $p$ signifies the probability of a sequence being a text modality. Next, we set a threshold $\tau$: when $p<\tau$, we perform mix-up on the ST sequence, while for $p>\tau$, we work with the MT sequence.

During mix-up of the ST representation processed by A-enc, each sequence position is replaced with the MT representation using probability $p$. The highest-probability position is determined through CTC prediction, and the ST feature is substituted with the MT embedding. Conversely, for the MT task, identifying an appropriate ST representation for mix-up proves challenging. To address this, we introduce audio-like noise (e.g., blank tokens, repeated adjacent tokens) to the MT sequence with probability $1-p$. Post mix-up, the target label of $\mathcal{D}$ becomes $p$ which is aligned with $U(0,1)$. By further sampling $p$ during training, the prediction space transforms from $\{c_\mathrm{st}, c_\mathrm{mt}\}$ to $[c_\mathrm{st}, c_\mathrm{mt}]$. Note that the mix-up feature does not feed into the decoder, thus the model will not directly benefits from the mix-up method.

\section{Experiments}
\label{sec:exprtiments}

\subsection{Datasets}

We conducted our experiments on the MuST-C English-German (En-De), English-French (En-Fr) and English-Spanish (En-Es) corpora \cite{di-gangi-etal-2019-must}. MuST-C is a multilingual speech translation corpus that comprises speeches, transcripts and translations from TED Talks. We utilized the WMT16, WMT14 and WMT13 for En-De, En-Fr and En-Es three tasks respectively, as the external datasets for pre-training our textual encoder and decoder. We implemented SentencePiece \cite{kudo2018sentencepiece} segmentation with a vocabulary size of 10,000 to build the shared vocabulary for every dataset.
\subsection{Experimental Settings}

We used Fairseq toolkit \cite{ott2019fairseq, wang-etal-2020-fairseq} to build all the models with about 150M parameters. We initialized our acoustic encoder by HuBert-base \cite{hsu2021hubert} which is a self-supervised pre-trained model. The textual encoder and decoder contain 6 layers which are initialized by the pre-trained MT model. The MT training process was followed by Ye~\etal~\cite{ye-etal-2022-cross}. For the adversarial training, the modal network consists of three feed-forward network layers and one output layer to reduce the feature dimension. We set the hidden size to 512. We used the pre-trained MT model and removed alignment methods as the single model of MT and ST tasks respectively.

For the multi-task learning process, we set the training step to 50k. The speech input is the raw audio and batch-size is less 16M audio frames. The learning rate is 2e-4. Regarding the initial weights, we set both the ASR and ST task weights to 1.0, while the MT task weight is set to 0.5. We set AT loss weight $\lambda$ to 3.5. During training, we limited the updates for the ASR task to 15000 steps. We set the $\tau$ to 0.1 to avoid injecting too much noise into the text. During inference, we averaged the best 5 checkpoints for evaluation. We use the beam size 8 and report sacreBLEU \cite{post-2018-call} for MT and ST tasks. All experiments were conducted on 8 NVIDIA 3090 GPUs.  
\subsection{Results}

Table \ref{results} presents the ST and MT results for three language tasks in the MuST-C dataset. Both S-Align and H-Align methods significantly improve ST performance over the single model. However, the results diverge when examining MT performance. Utilizing H-Align leads to a decline in MT scores across all three languages, with a significant 9.6 BLEU drop in the En-Fr task with external MT data. Interestingly, the En-Es task did not exhibit a significant MT performance decrement when H-Align was applied, although the gains in the ST task were marginal compared to those achieved by S-Align. These observations suggest that H-Align appears to be a trade-off strategy, affecting one task while benefiting another. Notably, our proposed S-Align method achieves MT performance on par with that of a dedicated MT model, thereby validating its capability to address both ST and MT tasks effectively. This supports our original hypothesis that S-Align can align the feature spaces of both modalities without causing adverse effects.

\section{Analysis}
\label{sec:analysis}

\subsection{Ablation Study}

We performed ablation studies on the MuST-C En-De and the results are displayed in Table \ref{Ablation}. When we entirely remove S-Align, the performance of both tasks experiences significant deterioration. This underscores the essential nature of the alignment method in bridging the modality gap. Moreover, when we apply enhanced adversarial training, both ST and MT show improvements, indicating that perturbing the representation is an effective strategy.

\begin{table}[t] 
\centering
\small
\setlength{\tabcolsep}{3mm}{
\begin{tabular}{lcc}
\toprule
Models&ST&MT \\
\hline

\textbf{S-Align}& 28.6 & 33.8 \\
\quad$-$ Enhanced adversarial training& 28.5 & 33.4 \\
\quad$-$ Adversarial training& 27.9 & 32.4 \\
\bottomrule
\end{tabular}
}
\caption{Ablation results on MuST-C En-De with external MT data.} 
\label{Ablation}
\end{table}

\begin{figure} [t]
\centering  
\subfigure[H-Align]{
\includegraphics[width=0.42\columnwidth, trim=42pt 42pt 20pt 20pt, clip]{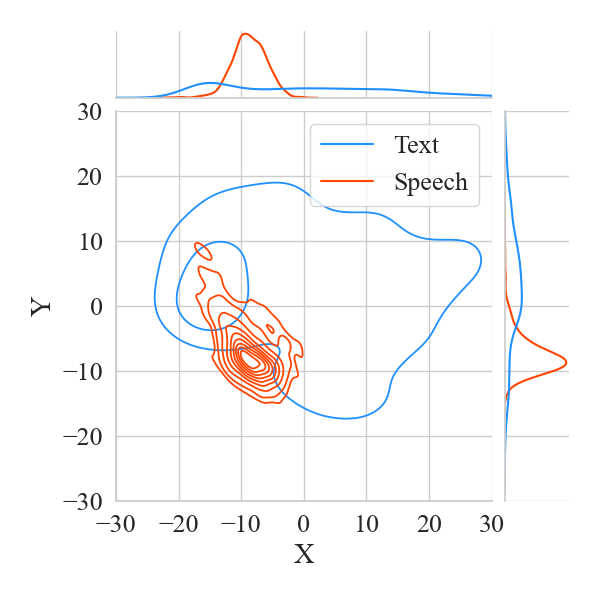}
}
\subfigure[S-Align]{
\includegraphics[width=0.42\columnwidth, trim=42pt 42pt 20pt 20pt, clip]{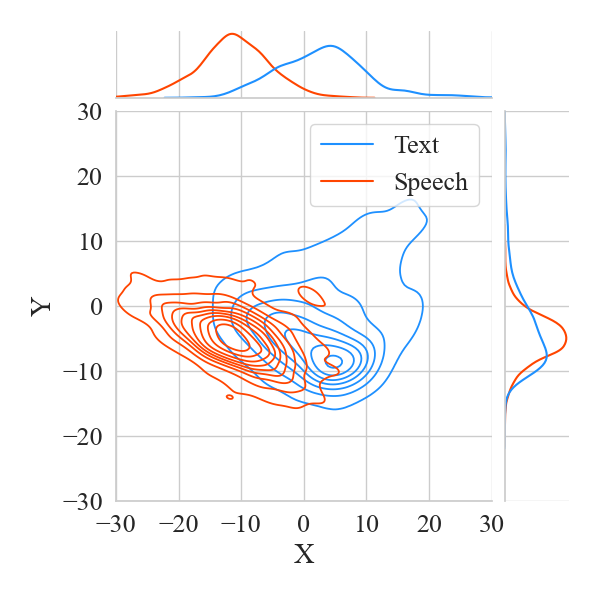}
}
\caption{Effects of alignment methods on modality space.}
\label{fig:alignment_effect}
\end{figure}
 
\subsection{Effect on Modality Space}

We extracted the two modality representations from the encoders on the En-De validation set and then applied Principal Component Analysis (PCA) to reduce the feature dimensions to two. Differing from the T-SNE \cite{JMLR:v9:vandermaaten08a}, the PCA tends to make the reduced point unique. The results presented in Figure \ref{fig:alignment_effect} demonstrate that both methods can achieve a certain alignment effect. The H-Align even brings the two modality representations closer together in terms of their center compared to S-Align. However, when considering the textual representation, it becomes evident that H-Align significantly disrupts the text space, causing it to become dispersed. In contrast, the S-Align method effectively preserves both modality spaces. This phenomenon confirms that S-Align can achieve a better alignment effect, whereas H-Align leads to a degradation in MT performance.

\subsection{Combination of Two Alignment Methods}

We combined S-Align and H-Align to assess their effects, as shown in Table \ref{Combine}. We tested contrastive learning methods as H-Align, which operate at two different levels \cite{ye-etal-2022-cross}. The \textit{high-level} refers to applying the alignment on semantic representations that are outputs from the textual encoder, while \textit{low-level} refers to the lexical embeddings before they are fed into the encoder. The results show that once we apply H-Align based on S-Align, the performance deteriorates. This confirms that alignment cannot always enhance performance due to the inherent differences between the two modalities.

\begin{table}[t] 
\centering
\small
\setlength{\tabcolsep}{3mm}{
\begin{tabular}{lcc}
\toprule
Models&ST&MT \\
\hline

\textbf{S-Align}& 28.6 & 33.8 \\
\quad$+$ low-level H-Align&28.4&33.7\\
\quad$+$ high-level H-Align& 28.5  &33.6   \\
\bottomrule
\end{tabular}
}
\caption{Combination of soft and hard alignment methods.}
\label{Combine}
\end{table}

\begin{figure}[t]
    \centering
        \begin{tikzpicture}
  
        \pgfplotsset{set layers}
         \scriptsize{
         \begin{axis}[at={(0,0)},
          ymajorgrids,
          xmajorgrids,
          grid style=dashed,
          width=.46\textwidth,
          height=.18\textwidth,
          legend style={at={(0.00,0.92)}, anchor=south west},
          xlabel={\scriptsize{Training Steps (K)}},
          ylabel={\scriptsize{Loss}},
          ylabel style={xshift=0em,yshift=-0.5em},
          yticklabel style={/pgf/number format/precision=3, /pgf/number format/fixed},
          ymin=0,ymax=4.2, ytick={0.5,1.5,2.5,3.5},
          xmin=19,xmax=41,xtick={20,25,30,35,40},
          legend columns=2,
          legend style={yshift=-10.5pt,xshift=0.5em, legend plot pos=right,cells={anchor=west},fill opacity =0.7}
          ]
          
          \addplot[orangered!60,mark=pentagon*,mark size=1.2pt,thick,mark options={fill=white,draw=orangered,line width=0.5pt}] coordinates {(20.6, 1.4102) (21.1, 1.4956) (21.6, 1.3127) (22.1, 1.4078) (22.6, 1.3892) (23.1, 1.4399) (23.6, 1.1654) (24.1, 1.1675) (24.6, 1.4868) (25.1, 1.3515) (25.6, 1.0096) (26.1, 0.9900) (26.6, 1.0268) (27.1, 1.1091) (27.6, 0.8896) (28.1, 0.5657) (28.6, 0.7448) (29.1, 0.6557) (29.6, 0.4884) (30.1, 1.7986) (30.6, 1.7302) (31.1, 1.2739) (31.6, 1.0839) (32.1, 0.7201) (32.6, 1.4381) (33.1, 1.8877) (33.6, 1.3488) (34.1, 0.9137) (34.6, 1.2356) (35.1, 0.6376) (35.6, 0.7011) (36.1, 0.6931) (36.6, 0.9371) (37.1, 0.3569) (37.6, 1.6496) (38.1, 1.2666) (38.6, 0.8676) (39.1, 0.7048) (39.6, 0.8372)}; 
          \addlegendentry{\scalebox{.8}{Discriminator}}
          \addplot[dodgerblue!60,mark=square*,mark size=1.0pt,thick,mark options={fill=white,draw=dodgerblue,line width=0.5pt}] coordinates {(20.6, 1.7683) (21.1, 1.8266) (21.6, 1.7902) (22.1, 1.7697) (22.6, 1.7929) (23.1, 1.8371) (23.6, 1.8609) (24.1, 1.9665) (24.6, 1.9198) (25.1, 1.8566) (25.6, 2.0070) (26.1, 2.1721) (26.6, 2.2191) (27.1, 2.0390) (27.6, 2.4995) (28.1, 2.9025) (28.6, 2.8639) (29.1, 2.9719) (29.6, 3.1263) (30.1, 2.2894) (30.6, 2.0479) (31.1, 1.9583) (31.6, 2.0904) (32.1, 2.5458) (32.6, 2.5380) (33.1, 2.0475) (33.6, 1.9990) (34.1, 2.0158) (34.6, 2.2202) (35.1, 2.8458) (35.6, 2.9975) (36.1, 2.8701) (36.6, 2.9611) (37.1, 3.6791) (37.6, 2.9284) (38.1, 2.5611) (38.6, 2.6888) (39.1, 3.0688) (39.6, 3.1484)};
          \addlegendentry{\scalebox{.8}{Generators}}
          \end{axis}}

        \end{tikzpicture}
    \caption{Change of adversarial training losses along training steps.}
    \label{at_loss}
\end{figure}
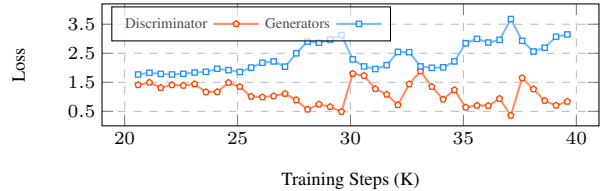

\subsection{Training Curves of Soft Alignment}

We display the training losses of the two generators and the discriminator in Figure \ref{at_loss}. \textit{Generators} legend denotes the sum of the losses from both generators. We observe that the two types of losses reverse throughout the entire training process, confirming the effectiveness of the adversarial strategy. In the later stages of training, both losses continue to increase slowly, indicating that the discriminator get harder to classify between the two modalities. This aligns with our goal of maintaining certain differences between the two spaces while ensuring that both tasks can be effectively addressed. 

\begin{table}[h!]
\centering
\small
\setlength{\tabcolsep}{3mm}{
\begin{tabular}{lll} 
\toprule
Models&ASR (WER$\downarrow$) \\
\hline
Single model& 14.73 \\
S-Align& 14.87 (+0.14)\\
H-Align& 15.85 (+1.12)\\
Combination & 15.13 (+0.40) \\
\bottomrule
\end{tabular}
}
\caption{ASR results on MuST-C En-De. The decoding process is non-autoregressive with CTC \cite{lee2021intermediate}.} 
\label{asr_results}
\end{table}

\subsection{Performance of ASR Task}

We present the ASR performances in Table \ref{asr_results} to assess the impact of the alignment method on acoustic features. Please note that we use the source text from the test set as the transcription, which includes punctuation. We observe that S-Align makes a minor impact on the acoustic encoder and achieves comparable performance to the single ASR model. However, when H-Align is applied, the performance decreases by 1.12 WER. Even when we combine the two methods, the decrease in performance persists. This demonstrates that H-Align negatively affects the low-level speech representation, although the output of T-enc benefits from H-Align.

\section{Conclusion}
\label{sec:conclusion}
The end-to-end ST model faces a modality gap issue. Previous work primarily concentrated on hard alignment methods to tackle this challenge. In contrast, we employ soft alignment through adversarial training to bridge the gap between the two representations. Our experiments demonstrate that soft alignment outperforms hard alignment by achieving modality alignment without compromising any representation space. Additionally, our model effectively handles MT, ST, and ASR tasks simultaneously, leveraging this advantage. 

\section{ACKNOWLEDGEMENT}
This work was supported in part by the National Science Foundation of China (No.62276056), the National Key R\&D Program of China, the Natural Science Foundation of Liaoning Province of China (2022-KF-16-01), the Fundamental Research Funds for the Central Universities (Nos. N2216016, N2216001, and N2216002), and the Program of Introducing Talents of Discipline to Universities, Plan 111 (No.B16009). The authors would like to thank anonymous reviewers for their insightful comments.

\bibliographystyle{IEEEbib}
\bibliography{refs}
\end{document}